\documentclass{article}
\pdfoutput=1 
\usepackage{fancyhdr}
\usepackage{graphicx}
\usepackage{amsmath}
\usepackage{amssymb}
\usepackage{authblk}
\usepackage{hyperref}
\usepackage{geometry}
\usepackage{pgfplots}
\usepgfplotslibrary{groupplots}

\title{Interpretable Multivariate Time Series Forecasting Using Neural Fourier Transform}

\author[1]{Noam Koren}
\author[2]{Kira Radinsky}
\affil[1,2]{Technion - Israel Institute of Technology, Haifa, Israel}
\affil[1]{\texttt{noam.koren@campus.technion.ac.il}}
\affil[2]{\texttt{kiraradinsky@gmail.com}}

\begin{document}

\maketitle

\begin{abstract}
Multivariate time series forecasting is a pivotal task in several domains, including financial planning, medical diagnostics, and climate science. This paper presents the Neural Fourier Transform (NFT) algorithm, which combines multi-dimensional Fourier transforms with Temporal Convolutional Network layers to improve both the accuracy and interpretability of forecasts. The Neural Fourier Transform is empirically validated on fourteen diverse datasets, showing superior performance across multiple forecasting horizons and lookbacks, setting new benchmarks in the field. This work advances multivariate time series forecasting by providing a model that is both interpretable and highly predictive, making it a valuable tool for both practitioners and researchers. The code for this study is publicly available\footnote{\url{https://github.com/2noamk/NFT}}.
\end{abstract}

\section{Introduction}
\label{introduction}
Time series forecasting, the process of predicting future values based on observed historical data, is a pivotal task in various fields such as economics, finance, medicine, and environmental science. This task becomes particularly complex when dealing with multivariate temporal data, where predicting future values involves understanding the intricate interdependencies among multiple variables. This is essential in scenarios like weather forecasting, where variables such as temperature and humidity are interlinked, or in financial markets, where the stock prices of interconnected companies are observed.

Recent advancements in time series forecasting have transitioned from conventional statistical approaches \cite{lutkepohl2005new, pfeffermann1989multivariate} to sophisticated machine learning techniques, notably deep learning \cite{lea2016temporal, patchtst, oreshkin2019n}. However, the field still grapples with a scarcity of both precise and interpretable models for multivariate time series prediction. The extension of interpretable methods from univariate to multivariate scenarios entails more than an increase in input dimensions; it demands a complex redesign to capture the interdependencies among multiple variables effectively.

In response to this challenge, we introduce a new architecture for multivariate predictions, termed the Neural Fourier Transform (NFT). Fourier Transforms, known for their mathematical simplicity and interpretability, surpass many deep learning models in these aspects. They simplify the analysis of time series by decomposing them into frequency components, thus aiding in the identification and understanding of periodic patterns and trends.
Moreover, Fourier Transforms are particularly adept at processing time series with pronounced periodic elements. Their ability to discern and isolate periodic signals renders them suitable for tasks in signal processing and seasonal pattern analysis. Furthermore, in instances involving noisy time series data, Fourier Transforms demonstrate greater robustness relative to deep learning models by efficiently segregating and eliminating noise, enhancing the overall prediction quality.

The Neural Fourier Transform (NFT) is an innovative approach that merges multi-dimensional Fourier transforms with Temporal Convolutional Network (TCN) layers \cite{lea2016temporal}, creating an explicit representation of seasonality in time series data. It excels in identifying seasonal components in multivariate time series by considering both the seasonal fluctuations within individual time series and the complex interrelations among different variables. This holistic perspective yields a more precise and detailed understanding of seasonal patterns within multivariate datasets.

In detail, the NFT utilizes an inverse multi-dimensional Discrete Fourier Transform (MFT) \cite{tolimieri2012mathematics}, implemented via two separate 1-DFT operations: one along the time axis and another across the variables. This dual application permits a thorough exploration of the time series, capturing critical patterns in both the temporal and variable dimensions.
Additionally, the model employs Temporal Convolutional Network (TCN) layers to learn the Fourier coefficients, thereby enhancing its capacity to detect complex temporal dependencies. As a block-based model, the NFT forecasts by discerning trends and seasonal shifts in time series data, achieving not only superior performance in multivariate time series forecasting compared to existing state-of-the-art (SOTA) methods but also providing crucial interpretability. This interpretability is particularly valuable for analyzing complex datasets, bridging the gap between advanced forecasting techniques and the need for model transparency. 
Moreover, the NFT model is designed for efficiency, completing its computations rapidly and enabling its use in real-time applications.

NFT is empirically evaluated using fourteen diverse datasets: Electricity consumption, weekly recorded influenza-like illness (ILI), daily Exchange Rates, hourly Traffic data, four ETT datasets (ETTh1, ETTh2, ETTm1, ETTm2) containing data
collected from electricity transformers, 12-lead Electrocardiogram (ECG), 12-lead ECG of 600 patients, 36-lead Electroencephalogram (EEG), Air Quality data from an Italian city, Chorales composed by Johann Sebastian Bach, and Weather records from The Global Historical Climatology Network daily (GHCNd). The results demonstrate significant improvements across multiple prediction horizons and lookbacks, confirming the efficacy of the proposed model.

The contribution of this work is threefold: 
First, The Neural Fourier Transform (NFT) is introduced, merging multi-dimensional Fourier transforms with Temporal Convolutional Networks reaching SOTA results for multivariate time series forecasting. 
Second, the study focuses on the interpretability of NFT, clearly representing seasonal and trend patterns in complex data. 
Last, the NFT's accuracy and interpretability are empirically validated across fourteen diverse datasets, outperforming current SOTA methods.

\section{Related Work}
\label{relatedWork}

\subsection{Interpretable Time Series Forecasting}
Machine learning approaches for interpretable time series forecasting have primarily been developed for univariate series, leaving a significant gap in methodologies suited for multivariate series. 

One example of an interpretable univariate algorithm is the Prophet algorithm \cite{zunic2020application} that gained traction for its adaptability in interpretable forecasting \textit{univariate} time series data. Building on the foundation laid by Prophet, Neural Prophet \cite{triebe2021neuralprophet} integrates neural network components with traditional forecasting techniques.

Recently, the N-BEATS algorithm \cite{oreshkin2019n} emerged as a notable neural-network-based model for \textit{univariate} time series forecasting. It has gained recognition for its balance of interpretability and efficiency, as evidenced by its top performance in the M5 Forecasting Competition \cite{makridakis2022m5}. N-BEATS has also been effectively applied in various fields, including Energy Forecasting \cite{oreshkin2021n}, Financial Market Forecasting \cite{liu2023s}, and Healthcare \cite{jossou2022n}.
N-BEATS features specific blocks for distinct patterns: Trend, Seasonality, and Generic. In the Trend Block, fully connected layers learn polynomial coefficients to model long-term data trends. The Seasonality Block captures periodic patterns by learning Fourier coefficients through its fully connected layers.
Complementing these two is the Generic Block, this block comprises fully connected layers to capture a wide array of patterns.
However, the N-BEATS model lacks the ability to predict multivariate data. Adapting it for multivariate forecasting involves more than just expanding the input dimensionality, it necessitates a deep understanding of the intricacies among multiple time series to effectively capture the interdependencies between variables.

\subsection{Multidimensional Fourier Transform}
The Fourier Transform has been widely used in time series analysis to identify periodic patterns or cycles in the data \cite{yi2023survey}. By converting time series data into the frequency domain, one can identify the main frequencies at which these cycles occur \cite{nussbaumer1982fast}.

Autoformer employs Fast Fourier Transforms (FFT) to compute autocorrelations efficiently \cite{wu2021autoformer}. FEDformer uses Fourier Transforms in its "Fourier-enhanced Federated Attention" to improve time series forecasting by concentrating on key frequencies \cite{zhou2022fedformer}. The Fourier Neural Operator (FNO) applies Fourier Transforms to effectively approximate operators in solving high-dimensional partial differential equations (PDEs) \cite{fno}. TimesNet utilizes Fourier Transforms for feature decomposition, which enables the analysis of time series in the frequency domain and the capture of essential periodic patterns, thus enhancing its forecasting capabilities \cite{wu2022timesnet}.

The \emph{Multidimensional Fourier Transform (MFT)} \cite{tolimieri2012mathematics} extends the capabilities of the traditional Fourier Transform to handle multi-dimensional data. While it has found widespread applications in fields such as image processing, where it aids in mathematical transformations involving translation, rotation, and scaling \cite{rao2021global}, \cite{soon2003speech}, \cite{gelman2007new}, its exploration in the time series domain remains relatively unresearched. This presented us with an opportunity for novel applications and advancements in analyzing multidimensional time series data, potentially unlocking new insights and methodologies in this area.

\section{Problem Statement}
\label{sec:problem_statement}

In this study, we address the multivariate point forecasting problem in discrete time. We are given a series history of length $T$, $\mathbf{y} = [y_1, \ldots, y_T] \in \mathbb{R}^{M \times T}$, and our task is to predict the matrix of future values $\mathbf{Y} = [y_{T+1}, \ldots, y_{T+H}] \in \mathbb{R}^{M \times H} $, where $y_t \in \mathbb{R}^M$ for each $t = 1, \ldots, T+H$, $M$ is the number of variables, and $H$ is the forecast horizon.

To simplify, we consider a lookback window of length $t \leq T$, ending at the most recent observation $y_T$. This window serves as the input to our model and is denoted by $\mathbf{X} \in \mathbb{R}^{M \times t} = [y_{T-t+1}, \ldots, y_T]$. The forecast of $\mathbf{Y}$ is represented as $\hat{\mathbf{Y}}$.

\section{Neural Fourier Transform (NFT)}
\begin{figure}[ht]
\centering
\includegraphics[width=0.45\textwidth]{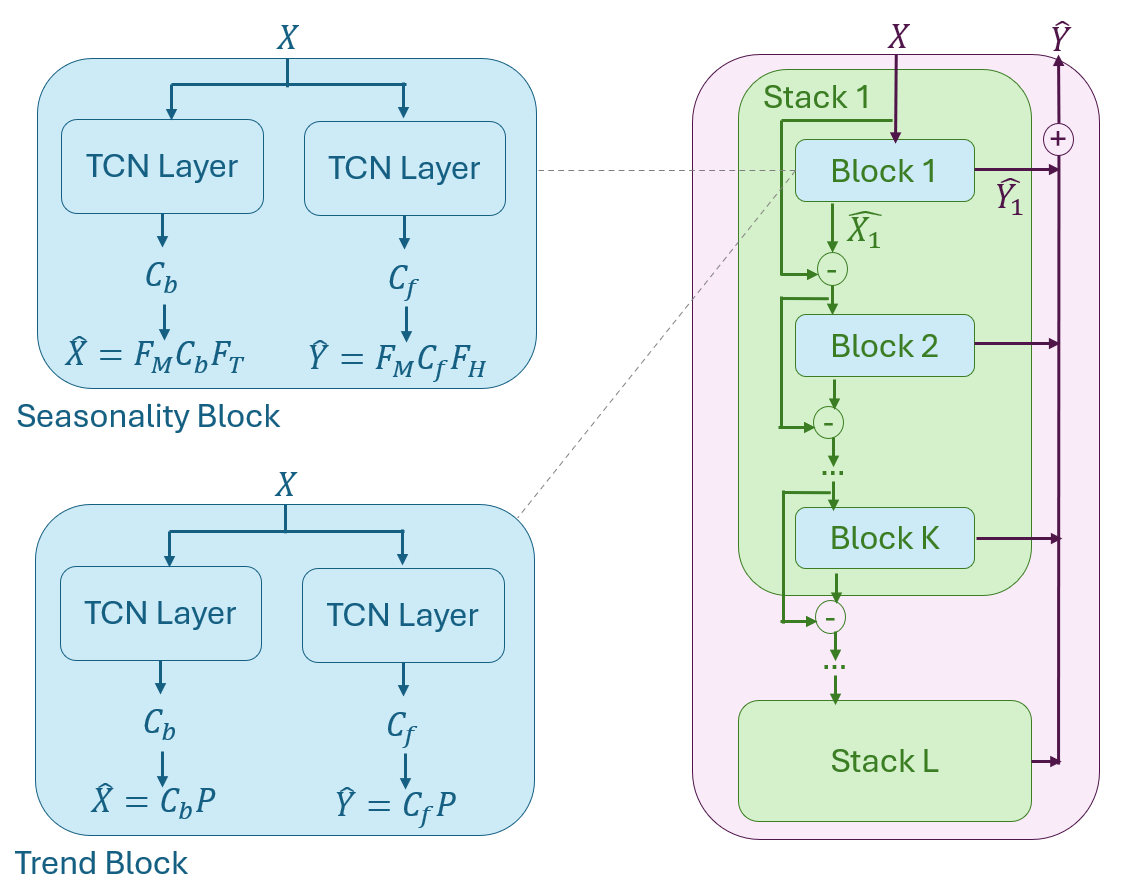}
\caption{Architecture of the Neural Fourier Transform (NFT) model. The diagram illustrates the arrangement of the Seasonality and Trend Blocks within the model, highlighting the multi-dimensional Fourier Transform and Temporal Convolutional Network layers.}.

\label{fig:NFT_architecture}
\end{figure}

In this section we present the Neural Fourier Transform, or in short NFT. The architecture of the NFT model is depicted in Figure \ref{fig:NFT_architecture}. Our algorithm identifies seasonality within the multivariate temporal data by employing the 2-dimensional DFT (2-DFT) \cite{tolimieri2012mathematics}. This technique decomposes the dataset into its core frequency components, both across the spectrum of variables and throughout the timeline. This becomes crucial when dealing with datasets where the interaction between different variables may give rise to new seasonality patterns that are not immediately observable in a univariate context.

Let's consider how 2-DFT is applied to the matrix $\mathbf{Y} \in \mathbb{R}^{M \times H}$.
The 2-DFT allows for a breakdown into two sequential 1-DFTs. The first stage applies a 1-DFT to columns of \(Y\), yielding an intermediate matrix \(Z\):
\begin{equation}
Z = F_M Y
\end{equation}
The subsequent phase involves a row-wise 1-DFT on \(Z\), deriving the matrix \(C\):
\begin{equation}
C = ZF_H^\top
\end{equation}
Concisely, this can be denoted as:
\begin{equation}
\label{eq:Fourier Coeffs mat}
C = F_M Y F_H^\top
\end{equation}
In this matrix representation, \(C\) contains the Fourier coefficients.

To revert from the frequency domain, and approximate the original matrix, the inverse operation is:
\begin{equation}
\label{eq:Fourier mat rep}
\hat{Y} = F_M^\top \times C \times F_H
\end{equation}

The NFT's primary objective is to learn the Fourier coefficients for the inverse 2-DFT from the matrix $\mathbf{X} \in \mathbb{R}^{M \times T}$. 
Note that these coefficients (\(C\)) depend on the unknown forecast \(Y\), and hence have to be learned by NFT.
The learned coefficients, \(C\), are then transformed from the frequency domain to the time domain. This transformation is accomplished through an inverse 2-DFT, as outlined in equation \ref{eq:Fourier mat rep}, enabling the model to generate forecasts for the specified horizon. During the training phase, the model iteratively adjusts these coefficients in response to the computed loss, fine-tuning them to accurately capture the underlying seasonal trends.

NFT diverges from the conventional 2-DFT approach by utilizing specially tailored Fourier matrices for the transformation process. Unlike the standard square matrix typically used in Fourier analysis, the matrix \( F_H \) is not necessarily square. It represents the granularity of the Fourier series across the time dimension. The dimensions of this matrix, \( \textit{Fourier order} (N) \times \textit{horizon length} (H) \), allow for a more flexible adaptation to the specific characteristics of the time series data. These are the following Fourier matrices \( F_M\) and \( F_H \) that achieve the desired Fourier transformation:

\[
F_M = 
\begin{bmatrix}
\cos(2\pi \cdot 0 \cdot \frac{0}{M}) & \cdots & \cos(2\pi \cdot 0 \cdot \frac{M-1}{M}) \\
\vdots & \ddots & \vdots \\
\cos(2\pi \cdot \frac{M}{2} \cdot \frac{0}{M}) & \cdots & \cos(2\pi \cdot \frac{M}{2} \cdot \frac{M-1}{M}) \\
\sin(2\pi \cdot 0 \cdot \frac{0}{M}) & \cdots & \sin(2\pi \cdot 0 \cdot \frac{M-1}{M}) \\
\vdots & \ddots & \vdots \\
\sin(2\pi \cdot \frac{M}{2} \cdot \frac{0}{M}) & \cdots & \sin(2\pi \cdot \frac{M}{2} \cdot \frac{M-1}{M}) \\
\end{bmatrix}
\]

\[
F_H = 
\begin{bmatrix}
\cos(2\pi \cdot 0 \cdot \frac{0}{H}) & \cdots & \cos(2\pi \cdot 0 \cdot \frac{H-1}{H}) \\
\vdots & \ddots & \vdots \\
\cos(2\pi \cdot \frac{N}{2} \cdot \frac{0}{H}) & \cdots & \cos(2\pi \cdot \frac{N}{2} \cdot \frac{H-1}{H}) \\
\sin(2\pi \cdot 0 \cdot \frac{0}{H}) & \cdots & \sin(2\pi \cdot 0 \cdot \frac{H-1}{H}) \\
\vdots & \ddots & \vdots \\
\sin(2\pi \cdot \frac{N}{2} \cdot \frac{0}{H}) & \cdots & \sin(2\pi \cdot \frac{N}{2} \cdot \frac{H-1}{H}) \\
\end{bmatrix}
\]

The matrix \( F_M \), corresponds to the number of variables in the dataset. It has a square dimension of \( \textit{number of variables} (M) \times \textit{number of variables} (M) \), reflecting the transformation applied across the different variables of the dataset. On the other hand, \( F_H\), relates to the temporal structure of the data. This matrix has dimensions of \( \textit{Fourier order} (N) \times \textit{horizon length} (H) \), where \textit{horizon length} (H) is the number of time points in each series and \textit{Fourier order} (N) represents the level of detail or granularity in the frequency domain, which we have selected based on the specific seasonal patterns we wish to capture.
Together, these matrices facilitate a comprehensive multidimensional Fourier analysis, enabling the decomposition of our multivariate time series into sinusoidal components.

\subsection{Learning Fourier Coefficients via TCN}
To address Fourier coefficient learning in multivariate time series, the NFT model employs Temporal Convolutional Network (TCN) layers. One could consider employing traditional Fully Connected (FC) layers to determine coefficients, e.g., as done by N-BEATS. However, the added complexity of multivariate time series, characterized by an additional dimension due to multiple variables, necessitates a more evolved approach that fits the nature of such data. A conventional strategy would transform the multivariate data from a three-dimensional structure \((\textit{batch size} \times M \times T)\)  into a two-dimensional format by flattening. Such a method does not align with the structure of the data and could potentially overlook the critical inter-variable correlations inherent in the data, besides imposing additional computational overhead.

To circumvent these limitations and to more effectively capture the complex interplay between variables, we propose the integration of Temporal Convolutional Networks (TCNs) in place of the Fully Connected layers. TCNs are inherently suited for multivariate time series as they are designed to preserve the data's original structural integrity, thereby avoiding the need for reshaping the data. 

\subsection{Expending NFT to Non-Seasonal Effects}
The Neural Fourier Transform captures seasonal patterns within a time series. However, time series are also compounded by a trend effect. 
Therefore, 
we extended the NFT model structure with specialized blocks tailored for different patterns. In addition to the Seasonality block, we add a Trend block and an optional Generic block. inspired by the N-BEATS framework, the NFT model arranges these blocks in a stack, where each block communicates and contributes to the next, ensuring a cohesive and integrated approach to modeling both seasonal and non-seasonal elements within the time series.

\subsubsection{Trend Block}
In the Trend Block, a polynomial-based approach is employed, tailored to learn the unique coefficients of a low-degree polynomial for each variable in the dataset.

Consider a multivariate time series \( \mathbf{Y} \) with \( H \) time points and \( M \) variables. The Trend Block approximates each variable's series using a polynomial of degree \( d \). This approximation can be represented as:
\begin{equation}
\hat{Y} = A \times P
\end{equation}
where \(A\) is the coefficients matrix, with dimensions \( M \times d \), and \(P\) is the Vandermonde matrix \cite{klinger1967vandermonde} constructed from the time vector \( t = [0, 1, ..., H-1]/H \), containing powers of \( t \) up to \( d-1 \), thus having dimensions \( d \times H \).

To effectively learn these coefficients, the Trend Block utilizes Temporal Convolutional Network (TCN) layers. As mentioned, TCNs are particularly adept at handling the temporal dependencies and complexities inherent in multivariate time series data, thus enabling a more accurate and tailored learning of polynomial coefficients for each variable. The learning process involves adjusting these coefficients to minimize the difference between the polynomial approximation \( \hat{Y}\) and the actual time series data \( Y\) for each variable.

\subsubsection{Generic Block}
The Generic Block is composed of Temporal Convolutional Network (TCN) layers which preserve the original multidimensional structure of the data and manage the complexities inherent in multivariate time series data. 

In this work, the generic block wasn't found useful, but in other setups, we envision it might find additional value.

\subsection{Model Interpretability}
An inherent advantage of the Neural Fourier Transform model is its interpretability, deriving from its capacity to learn explicit representations of the time series components. By directly learning the polynomial and Fourier coefficients, the model provides insights into the underlying trend and seasonality mechanisms of the series.
Within the trend blocks, the learned polynomial coefficients provide a mathematical insight into the series' long-term movements, helping to elucidate the nature and degree of trends, whether linear, quadratic, or more complex. Similarly, the Fourier coefficients within the seasonality block capture the cyclical patterns inherent in the data. These coefficients can be interpreted to discern the amplitude, frequency, and phase of the seasonal fluctuations, providing a transparent view of the periodicities that drive the multivariate time series.

\section{Empirical Evaluation}
\label{Evaluation}

\subsection{Experimental Methodology}
Two experimental protocols were employed:
\begin{enumerate} 
    \item \textit{\textbf{EvalProtocol1}}: This protocol focuses on predicting future time steps of a single multivariate time series. For instance, in the ECG dataset, the evolution of a single patient's ECG over time is forecasted.
    
    \item \textit{\textbf{EvalProtocol2}}: This protocol involves learning from several multivariate time series and evaluating the model on different series within the dataset. For example, in the ECG dataset, the model is trained on a subset of patients and tested on unseen patients. 
\end{enumerate}

The selection of protocols depended on the specific research objectives. \textit{EvalProtocol1} was ideal for tasks like climate prediction or personalized medical forecasts, where data specific to a single location or individual is analyzed. In contrast, \textit{EvalProtocol2} was better suited for situations lacking individual records, such as medical forecasts when patients' histories are unavailable or musical compositions.

\subsection{Datasets}
The experiments were conducted on fourteen diverse datasets:
\begin{itemize}
    \item Hourly Electricity consumption of 321 customers from 2012 to 2014\footnote{\url{https://archive.ics.uci.edu/dataset/321/electricityloaddiagrams20112014}}
    \item Weekly data on influenza-like illness (ILI) patients recorded by the Centers for Disease Control and Prevention of the United States from 2002 to 2021, detailing the proportion of patients presenting with ILI relative to the total patient count\footnote{\url{https://gis.cdc.gov/grasp/fluview/fluportaldashboard.html}}
    \item Daily Exchange Rates of eight different countries ranging from 1990 to 2016 \cite{wu2021autoformer}
    \item Hourly traffic data provided by the California Department of Transportation, detailing road occupancy rates recorded by various sensors located on freeways in the San Francisco Bay area\footnote{\url{https://pems.dot.ca.gov/}}
    \item Four ETT datasets (ETTh1, ETTh2, ETTm1, ETTm2) comprising data gathered from electric transformers, including load and oil temperature measurements taken every 15 minutes from July 2016 to July 2018 \cite{zhou2021informer}
    \item 12-lead Electrocardiogram (ECG) \cite{goldberger2000physiobank}
    \item 36-lead Electroencephalogram (EEG) \cite{obeid2016temple}
    \item Hourly Air Quality data from an Italian city from March 2004 to February 2005, recorded by multisensor devices \cite{air_Quality}
    \item Chorales composed by Johann Sebastian Bach \cite{Ottmanbach}
    \item Weather dataset includes daily climate summaries from three global stations: Sharjah, Dubai, and Abu Dhabi International Airports \footnote{\url{https://www.ncei.noaa.gov/products/land-based-station/global-historical-climatology-network-daily}}
\end{itemize}
Statistics on the datasets can be found in Table \ref{fig:datasets}.


The 12-lead ECG, 36-lead EEG, Weather, and Chorales datasets include multiple multivariate time series.
For instance, the 12-lead ECG dataset contains numerous patients, each represented as a 12-variable time series.

Data preprocessing included outlier removal using the Interquartile Range (IQR) method, imputation of missing values via mean substitution, and data standardization.

In our analysis, we applied the \textit{EvalProtocol1} approach to the Electricity, ILI, Traffic, ETT, and Exchange Rate datasets, allocating the initial 70\% of the data for training, the subsequent 10\% for validation, and the remaining 20\% for testing. This protocol was also used for the 12-lead ECG, 36-lead EEG, Air Quality, and Weather datasets, with different data partitioning: 50\% for training, 20\% for validation, and 30\% for testing. For the ECG and EEG datasets, we focused on forecasting for four and two specific individuals respectively, and averaged the results.

Additionally, we utilized the \textit{EvalProtocol2} approach for the 12-Lead ECG and Chorales datasets, which were comprised of data from 600 patients and denoted for the 12-lead ECG and 100 melodies for the chorales, dividing the data into 50\% for training, 15\% for validation, and 35\% for testing.

\begin{table*}[h!]
\centering
\caption{Datasets Statistics}
\begin{tabular}{c|c c c c c c c c c c}
\hline
Datasets     & ETTh1   & ETTh2       & ETTm1  & ETTm2      & ECG (600 patients) & EEG    & Chorales \\ \hline
Features     & 7       & 7           & 7      & 7          & 12                 & 36     & 6        \\ \hline
Timesteps    & 14592   & 14592       & 57792  & 57792      & 2393563            & 176395 & 4647     \\ \hline
Lookback     & 96      & 96          & 96     & 96         & 100                & 100    & 10       \\ \hline
Prediction   & 1, 10,  & 1, 10,      & 1, 10, & 1, 10,     & 1, 5,              & 1, 10, & 1, 2,    \\
Horizons     & 24, 48, & 24, 48,     & 24, 48,& 24, 48     & 10, 15,            & 25     & 3, 4,    \\
             & 96, 192,& 96, 192,    & 96, 192,& 96, 192,  & 20, 25,            &        & 5        \\
             & 336, 720& 336, 720    & 336, 720& 336, 720  & 30                 &        &          \\ \hline \hline
Datasets     & Traffic & Electricity & ILI & Exchange Rate & Weather & Air Quality & ECG     \\ \hline
Features     & 862     & 321         & 7   & 8             & 3       & 13          & 12      \\ \hline
Timesteps    & 17544   & 6304        & 966 & 7588          & 17235   & 9357        & 3989    \\ \hline
Lookback     & 96      & 96          & 36  & 96            & 360     & 40          & 200     \\ \hline
Prediction   & 1, 16,  & 1, 16,      & 1, 12, & 1, 16,     & 1, 7,   & 1, 5,       & 1, 10,  \\
Horizons     & 32, 48  &  32         & 24 ,36,& 32, 48,    & 15, 30, & 10, 15,     & 25, 50, \\
             &         &             & 48, 60 & 96, 192,   & 60, 90, & 20, 25,     & 100     \\
             &         &             &        & 336, 720   & 120, 180& 30          &         \\ \hline
             
\hline
\end{tabular}
\label{fig:datasets}
\end{table*}

\subsection{Baseline Models}
We compared the Neural Fourier Transform (NFT) model and various state-of-the-art forecasting architectures, including TimesNet \cite{wu2022timesnet}, PatchTST \cite{patchtst} DLinear \cite{zeng2023dlinear}, Autoformer \cite{wu2021autoformer}, Temporal Convolutional Network (TCN) \cite{lea2016temporal}, and N-BEATS \cite{oreshkin2019n}. Each model was tested on the fourteen diverse datasets, across different lookback periods and forecasting horizons, to ensure a comprehensive and robust evaluation.

Across all models, the Mean Squared Error (MSE) was utilized during model training. Hyperparameter tuning was conducted with a separate validation dataset. 

In the NFT model, the seasonality blocks were set with a Fourier granularity of 8, and the trend blocks had a polynomial degree of 4. 
The NFT model consists of two distinct stacks: one capturing trend and the other capturing seasonality. The number of blocks in each stack differed across datasets: For Weather, ILI, Electricity, Exchange Rate, Traffic, ETT, and ECG (600 patients), it comprised 2 blocks per stack, whereas, for ECG, EEG, Air Quality, and Bach Chorales datasets, 3 blocks per stack were used. The specific number of blocks for each dataset was determined based on the validation set. 

For the baseline models, the optimal parameters differed through datasets. This variation allowed each model to tailor its approach to best capture the dynamics of the specific time series data it was forecasting, creating a strong baseline.

Notably, the N-BEATS model is designed for univariate time series forecasting. Therefore, each variable was forecasted independently and the metrics were subsequently averaged across all variables for a comprehensive evaluation.



\section{Empirical Results}
\subsection{Main Result}
\begin{figure*}[htbp]
\centering
\begin{tikzpicture}
  \begin{groupplot}[
    group style={
      group size=3 by 5, 
      horizontal sep=2cm,
      vertical sep=1.8cm, 
    },
    width=5cm,
    height=3.5cm,
    legend style={at={(0.5,-0.7)}, anchor=north, legend columns=-1} 
  ]

  \pgfplotsset{
    nft/.style={blue, solid},
    timesnet/.style={orange, densely dotted},
    patchtst/.style={green, densely dashdotted},
    dlinear/.style={purple, densely dashdotted},
    Autoformer/.style={black, densely dotted},
    tcn/.style={teal, densely dashed},
    nbeats/.style={red, densely dotted},
  }
\nextgroupplot[xlabel={Horizon}, ylabel={MSE}, title={\textbf{Exchange Rate}}]
\addplot[nft] coordinates {(1,0.0001) (16,0.01) (32,0.02) (48,0.02) (96, 0.02) (192, 0.03) (336, 0.04) (720, 0.05) };
\addplot[timesnet] coordinates {(1,0.05) (16,0.18) (32,0.15) (48,0.11) (96, 0.107) (192, 0.226) (336, 0.367) (720, 0.964)  };
\addplot[patchtst] coordinates {(1,0.01) (16,0.02) (32,0.03) (48,0.04) (96, 0.107) (192, 0.226) (336, 0.367) (720, 0.964)  };
\addplot[dlinear] coordinates {(1,0.01) (16,0.02) (32,0.03) (48,0.05) (96, 0.088) (192, 0.176) (336, 0.313) (720, 0.839) };
\addplot[Autoformer] coordinates {(1,0.04) (16,0.04) (32,0.06) (48,0.09) (96, 0.197) (192, 0.300) (336, 0.509) (720, 1.447) };
\addplot[tcn] coordinates {(1,0.2) (16,0.11) (32,0.04) (48,0.06) (96, 0.09) (192, 0.17) (336, 0.24) (720, 0.41) };
\addplot[nbeats] coordinates {(1,0.01) (16,0.03) (32,0.05) (48,0.12) (96, 0.17) (192, 0.19) (336, 0.21) (720, 0.25) };

\nextgroupplot[xlabel={Horizon}, ylabel={MSE}, title={\textbf{ILI}}]
\addplot[nft] coordinates {(1,0.05) (12,0.18) (24,0.25) (36, 0.31) (48, 0.35) (60, 0.42)};
\addplot[timesnet] coordinates {(1, 0.64) (12,2.67) (24, 2.317) (36, 1.972) (48, 2.238) (60, 2.027)};
\addplot[patchtst] coordinates {(1,0.24) (12,1.25) (24,1.319) (36, 1.579) (48, 1.553) (60, 1.470)};
\addplot[dlinear] coordinates {(1,2.32) (12,2.83) (24, 2.215) (36, 1.963) (48, 2.13) (60, 2.368)};
\addplot[Autoformer] coordinates {(1,0.84) (12,2.47) (24, 3.483) (36, 3.103) (48, 2.669) (60, 2.770)};
\addplot[tcn] coordinates {(1,0.86) (12,0.98) (24,0.95) (36, 0.94) (48, 1.01) (60, 1.03)};
\addplot[nbeats] coordinates {(1,0.54) (12,2.68) (24,4.09) (36,4.16) (48,4.33) (60,4.69)};
    
\nextgroupplot[xlabel={Horizon}, ylabel={MSE}, title={\textbf{Air Quality}}]
\addplot[nft] coordinates { (1, 0.13) (5, 0.21) (10, 0.28) (15, 0.31) (20, 0.35) (25, 0.38) (30, 0.4) };
\addplot[timesnet] coordinates { (1, 0.74) (5, 0.81) (10, 0.85) (15, 0.87) (20, 0.85) (25, 0.92) (30, 0.96) };
\addplot[patchtst] coordinates { (1, 0.22) (5, 0.58) (10, 0.68) (15, 0.72) (20, 0.75) (25, 0.78) (30, 0.83) };
\addplot[dlinear] coordinates { (1, 0.39) (5, 0.69) (10, 0.78) (15, 0.83) (20, 0.87) (25, 0.93) (30, 1.01) };
\addplot[Autoformer] coordinates { (1, 0.72) (5, 0.86) (10, 0.97) (15, 1.02) (20, 0.98) (25, 1.03) (30, 0.99) };
\addplot[tcn] coordinates { (1, 0.21) (5, 0.42) (10, 0.53) (15, 0.64) (20, 0.69) (25, 0.68) (30,0.69) };
\addplot[nbeats] coordinates { (1, 0.23) (5, 0.61) (10, 0.76) (15, 0.81) (20, 0.9) (25, 0.93) (30, 1) };

\nextgroupplot[xlabel={Horizon}, title={\textbf{ETTh1}}]
\addplot[nft] coordinates {(1,0.07) (10,0.14) (24,0.18) (48,0.25) (96,0.39) (192,0.43) (336,0.44) (720,0.46)};
\addplot[timesnet] coordinates {(1,0.15) (10,0.29) (24,0.33) (48,0.37) (96,0.38) (192,0.44) (336,0.49) (720,0.52)};
\addplot[patchtst] coordinates {(1,0.1) (10,0.25) (24,0.29) (48,0.34) (96,0.37) (192,0.41) (336,0.42) (720,0.45)};
\addplot[dlinear] coordinates {(1,0.12) (10,0.28) (24,0.31) (48,0.35) (96,0.37) (192,0.41) (336,0.44) (720,0.47)};
\addplot[Autoformer] coordinates {(1,0.17) (10,0.4) (24,0.4) (48,0.427) (96,0.43) (192,0.46) (336,0.49) (720,0.52)};
\addplot[tcn] coordinates {(1,0.14) (10,0.29) (24,0.43) (48,0.46) (96,0.46) (192,0.54) (336,0.53) (720,0.58)};

\nextgroupplot[xlabel={Horizon}, title={\textbf{ETTh2}}]
\addplot[nft] coordinates {(1,0.06) (10,0.1) (24,0.13) (48,0.16) (96,0.19) (192,0.23) (336,0.26) (720,0.34)};
\addplot[timesnet] coordinates {(1,0.1) (10,0.15) (24,0.2) (48,0.25) (96,0.34) (192,0.40) (336,0.45) (720,0.46)};
\addplot[patchtst] coordinates {(1,0.06) (10,0.12) (24,0.17) (48,0.22) (96,0.274) (192,0.34) (336,0.33) (720,0.38)};
\addplot[dlinear] coordinates {(1,0.11) (10,0.13) (24,0.18) (48,0.25) (96,0.289) (192,0.38) (336,0.45) (720,0.61)};
\addplot[Autoformer] coordinates {(1,0.12) (10,0.21) (24,0.26) (48,0.32) (96,0.332) (192,0.43) (336,0.48) (720,0.45)};
\addplot[tcn] coordinates {(1,0.06) (10,0.11) (24,0.15) (48,0.19) (96,0.24) (192,0.48) (336,0.54) (720,0.55)};

\nextgroupplot[xlabel={Horizon}, title={\textbf{ETTm1}}]
\addplot[nft] coordinates {(1,0.04) (10,0.07) (24,0.11) (48,0.14) (96,0.16) (192,0.24) (336,0.37) (720,0.43)};
\addplot[timesnet] coordinates {(1,0.05) (10,0.13) (24,0.20) (48,0.29) (96,0.34) (192,0.37) (336,0.41) (720,0.48)};
\addplot[patchtst] coordinates {(1,0.05) (10,0.12) (24,0.21) (48,0.28) (96,0.293) (192,0.33) (336,0.37) (720,0.42)};
\addplot[dlinear] coordinates {(1,0.05) (10,0.14) (24,0.24) (48,0.31) (96,0.299) (192,0.34) (336,0.37) (720,0.43)};
\addplot[Autoformer] coordinates {(1,0.09) (10,0.24) (24,0.4) (48,0.56) (96,0.51) (192,0.51) (336,0.51) (720,0.53)};
\addplot[tcn] coordinates {(1,0.04) (10,0.09) (24,0.14) (48,0.19) (96,0.23) (192,0.47) (336,0.45) (720,0.54)};

\nextgroupplot[xlabel={Horizon}, title={\textbf{ETTm2}}]
\addplot[nft] coordinates {(1,0.03) (10,0.05) (24,0.07) (48,0.1) (96,0.13) (192,0.17) (336,0.22) (720,0.26)};
\addplot[timesnet] coordinates {(1,0.04) (10,0.07) (24,0.11) (48,0.15) (96,0.19) (192,0.25) (336,0.32) (720,0.41)};
\addplot[patchtst] coordinates {(1,0.03) (10,0.07) (24,0.1) (48,0.14) (96,0.166) (192,0.22) (336,0.27) (720,0.36)};
\addplot[dlinear] coordinates {(1,0.03) (10,0.07) (24,0.11) (48,0.15) (96,0.167) (192,0.22) (336,0.28) (720,0.40)};
\addplot[Autoformer] coordinates {(1,0.07) (10,0.1) (24,0.14) (48,0.165) (96,0.205) (192,0.28) (336,0.34) (720,0.41)};
\addplot[tcn] coordinates {(1,0.03) (10,0.06) (24,0.09) (48,0.12) (96,0.16) (192,0.30) (336,0.34) (720,0.42)};

\nextgroupplot[xlabel={Horizon}, ylabel={MSE}, title={\textbf{Traffic}}]
\addplot[nft] coordinates {(1, 0.09) (16, 0.18) (32, 0.26) (48, 0.4)};
\addplot[timesnet] coordinates {(1, 0.53) (16, 0.58) (32, 0.62) (48, 0.62)};
\addplot[patchtst] coordinates {(1, 0.26) (16, 0.58) (32, 0.63) (48, 0.66)};
\addplot[dlinear] coordinates {(1, 0.4) (16, 0.63) (32, 0.67) (48, 0.72)};
\addplot[Autoformer] coordinates {(1, 0.5) (16, 0.56) (32, 0.67) (48, 0.65)};
\addplot[tcn] coordinates {(1, 0.1) (16, 0.46) (32, 0.49) (48, 0.51)};
\addplot[nbeats] coordinates {(1, 0.2) (16, 0.51) (32, 0.59) (48, 0.71)};

\nextgroupplot[xlabel={Horizon}, ylabel={MSE}, title={\textbf{Weather}}]
\addplot[nft] coordinates { (1, 0.19) (7, 0.24) (15, 0.24) (30, 0.25) (60, 0.26) (90, 0.27) (120, 0.28) (180, 0.27) };
\addplot[timesnet] coordinates { (1, 0.33) (7, 0.35) (15, 0.44) (30, 0.36) (60, 0.37) (90, 0.37) (120, 0.38) (180, 0.37) };
\addplot[patchtst] coordinates { (1, 0.34) (7, 0.36) (15, 0.37) (30, 0.38) (60, 0.40) (90, 0.42) (120, 0.42) (180, 0.43) };
\addplot[dlinear] coordinates { (1, 0.29) (7, 0.32) (15, 0.33) (30, 0.35) (60, 0.37) (90, 0.4) (120, 0.41) (180, 0.4) };
\addplot[Autoformer] coordinates { (1, 0.37) (7, 0.37) (15, 0.37) (30, 0.38) (60, 0.42) (90, 0.44) (120, 0.46) (180, 0.47) };
\addplot[tcn] coordinates { (1, 0.25) (7, 0.29) (15, 0.30) (30, 0.32) (60, 0.36) (90, 0.33) (120, 0.33) (180, 0.35) };
\addplot[nbeats] coordinates { (1, 0.31) (7, 0.33) (15, 0.35) (30, 0.38) (60, 0.40) (90, 0.42) (120, 0.43) (180, 0.44) };

\nextgroupplot[xlabel={Horizon}, ylabel={MSE}, title={\textbf{Electricity}}]
\addplot[nft] coordinates {(1,0.04) (16,0.08) (32,0.13)};
\addplot[timesnet] coordinates {(1,0.14) (16,0.16) (32,0.17) };
\addplot[patchtst] coordinates {(1,0.09) (16,0.2) (32,0.17) };
\addplot[dlinear] coordinates {(1,0.07) (16,0.16) (32,0.18)};
\addplot[Autoformer] coordinates{(1,0.2) (16,0.4) (32,0.6)};
\addplot[tcn] coordinates {(1,0.05) (16,0.32) (32,0.44)};
\addplot[nbeats] coordinates {(1,0.06) (16,0.13) (32,0.18)};

\nextgroupplot[xlabel={Horizon}, ylabel={MSE}, title={\textbf{ECG (600 patients)}}]
\addplot[nft] coordinates {(1,0.02) (5,0.08) (10,0.07) (15,0.1) (20,0.15) (25,0.15) (30,0.1)};
\addplot[timesnet] coordinates {  (1,0.02) (5,0.08) (10,0.08) (15,0.1) (20,0.15) (25,0.15) (30,0.17)};
\addplot[patchtst] coordinates {  (1,0.02) (5,0.05) (10,0.08) (15,0.1) (20,0.15) (25,0.15) (30,0.18)};
\addplot[dlinear] coordinates { (1,0.02) (5,0.06) (10,0.08) (15,0.11) (20,0.14) (25,0.16) (30,0.19)};
\addplot[Autoformer] coordinates {(1,0.5) (5,0.49) (10,0.38) (15,0.41) (20,0.46) (25,0.44) (30,0.45)};
\addplot[tcn] coordinates {(1,0.03) (5,0.07) (10,0.1) (15,0.12) (20,0.16) (25,0.18) (30,0.21) };
\addplot[nbeats] coordinates {(1,0.04) (5,0.08) (10,0.12) (15,0.16) (20,0.20) (25,0.22) (30,0.25) };

\nextgroupplot[xlabel={Horizon}, ylabel={MSE}, title={\textbf{ECG}}]
\addplot[nft] coordinates {(1,0.04) (10,0.08) (25,0.12) (50,0.14) (100,0.18) };
\addplot[timesnet] coordinates {(1,0.31) (10,0.49) (25,0.625) (50,1.16) (100,1.09)};
\addplot[patchtst] coordinates {(1,0.34) (10,0.57) (25,0.77) (50,0.94) (100,0.98)};
\addplot[dlinear] coordinates { (1,0.35) (10,0.46) (25,0.58) (50,0.66) (100,0.69) };
\addplot[Autoformer] coordinates {(1,1.62) (10,1.13) (25,1.5) (50,1.15) (100,1.26)};
\addplot[tcn] coordinates {(1,0.16) (10,0.42) (25,0.56) (50,0.70) (100,0.81)};
\addplot[nbeats] coordinates {(1,0.13) (10,0.29) (25,0.39) (50,0.49) (100,0.59)};

\nextgroupplot[xlabel={Horizon}, ylabel={MSE}, title={\textbf{EEG}}]
\addplot[nft] coordinates { (1, 0.03) (10, 0.1) (25, 0.16)};
\addplot[timesnet] coordinates { (1, 0.16) (10, 0.35) (25, 0.44)};
\addplot[patchtst] coordinates { (1, 0.15) (10, 0.345) (25, 0.42)};
\addplot[dlinear] coordinates { (1, 0.135) (10, 0.365) (25, 0.435)};
\addplot[Autoformer] coordinates { (1, 0.63) (10, 0.545) (25, 0.62)};
\addplot[tcn] coordinates { (1, 0.09) (10, 0.32) (25, 0.39)};
\addplot[nbeats] coordinates { (1, 0.11) (10, 0.32) (25, 0.43)};

\nextgroupplot[xlabel={Horizon}, ylabel={MSE}, title={\textbf{Chorales}}]
\addplot[nft] coordinates {(1,0.15) (2,0.18) (3,0.20) (4,0.23) (5,0.22)};
\addplot[timesnet] coordinates {(1,0.29) (2,0.3) (3,0.31) (4,0.31) (5,0.32)};
\addplot[patchtst] coordinates {(1,0.34) (2,0.27) (3,0.29) (4,0.30) (5,0.31)};
\addplot[dlinear] coordinates {(1,0.29) (2,0.29) (3,0.4) (4,0.37) (5,0.36)};
\addplot[Autoformer] coordinates {(1,0.3) (2,0.25) (3,0.25) (4,0.26) (5,0.24)};
\addplot[tcn] coordinates {(1,0.27) (2,0.63) (3,0.62) (4,0.37) (5,0.56)};
\addplot[nbeats] coordinates {(1,0.66) (2,0.25) (3,0.27) (4,0.28) (5,0.26)};

\addlegendentry{NFT}
\addlegendentry{TimesNet}
\addlegendentry{PatchTST}
\addlegendentry{DLinear}
\addlegendentry{Autoformer}
\addlegendentry{TCN}
\addlegendentry{N-BEATS}
\end{groupplot}
\end{tikzpicture}
\caption{MSE results across different forecasting methods and datasets}
\label{fig:mse}
\end{figure*}
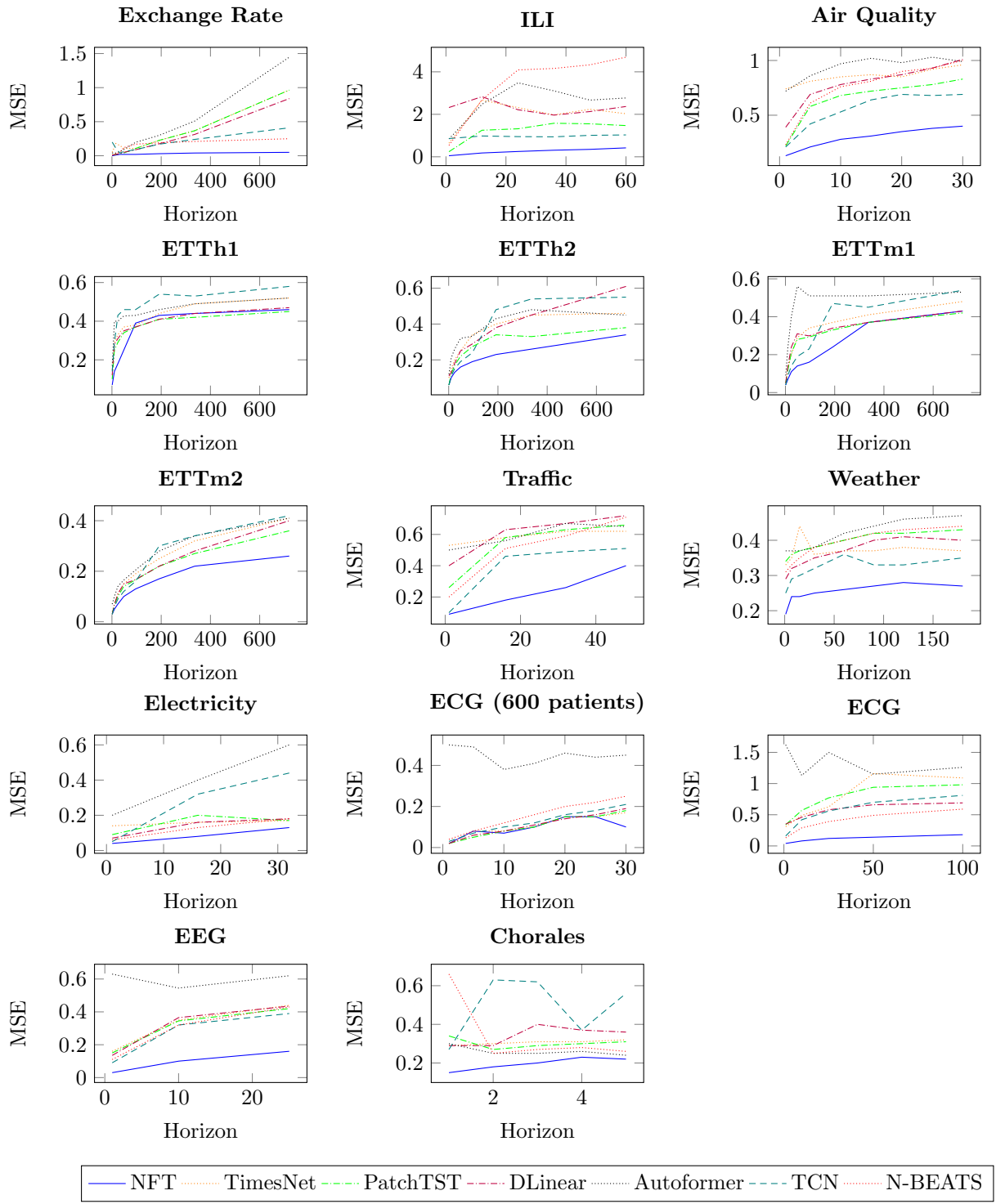

The results, displayed in Figure \ref{fig:mse}, highlight the NFT model's strong performance across a diverse range of datasets. Accuracy was measured via the mean squared error (MSE) metric, which is widely recognized in forecasting literature \cite{das2004mean, hyndman2006another, makridakis2000m3}. The model achieves SOTA performance and significant MSE reductions, with average improvement percentages of \textbf{82.69\%} in Exchange Rate, \textbf{71.01\%} in ILI, \textbf{46.04\%} in Air Quality, \textbf{15.14\%} in ETTh1, \textbf{15.39\%} in ETTh2, \textbf{15.66\%} in ETTm1, \textbf{17.91\%} in ETTm2, \textbf{34.84\%} in Traffic, \textbf{20.88\%} in Weather, \textbf{31.17\%} in Electricity, \textbf{77.44\%} in ECG, \textbf{5.53\%} in ECG of 600 patients, \textbf{64.80\%} in EEG,  and \textbf{22.46\%} in Chorales. For each dataset we averaged the MSE improvement percentages across time horizons. The improvement was calculated by comparing the NFT model to the best-performing baseline per each horizon.

To assess the NFT algorithm's robustness, we conducted paired $t$-tests across all datasets, yielding an average $t$-statistic of \textbf{7.21} and a $p$-value of \textbf{0.028}, confirming statistically significant performance over benchmark models.
Particularly, datasets such as Weather and Air Quality showed highly significant differences, with $p$-values of $6.84 \times 10^{-6}$ and $1.03 \times 10^{-4}$, respectively. 

Overall, the results indicate that the NFT model provides a robust solution for multivariate time series forecasting. Its consistent performance across different datasets, horizons, and lookbacks highlights its effectiveness for interpretable multivariate time series prediction.

\textbf{Speed and Efficiency} A standout feature of the NFT model is its rapid architecture, completing epochs in merely 26 seconds on average across all datasets, compared to the 150+ seconds required by TimesNet, PatchTST, Autoformer, and N-BEATS. 

\textbf{Performance Analysis by Prediction Range}
We analyzed the relative accuracy of NFT as a function of the prediction range. To this end, we computed, for each dataset, the correlation between the prediction range (number of time units into the future, previously denoted by $H$) and the improvement percentage (MSE-reduction relative to the closest baseline). See Table \ref{tab:correlations} for the resulting correlations. We observed a positive correlation for the datasets Traffic, Exchange Rate, ECG (600 patients), ETTh2, and ETTm2, suggesting that the NFT model increasingly improves compared to the baseline predictions as the prediction horizon grows. For the Air Quality, Electricity, and Weather datasets, the correlation was close to zero, indicating stable NFT improvement across different horizons. Conversely, for the ILI, ECG, EEG, Chorales, ETTh1, and ETTm1 datasets, we found a negative correlation, which may imply diminishing stability of the learned Fourier coefficients as the prediction horizon extends for these datasets.

\begin{table}[h]
\centering
\caption{Correlation between Prediction Ranges and Improvement Percentages}
\label{tab:correlations}
\begin{tabular}{|l|r|}
\hline
\textbf{Dataset}         & \textbf{Correlation} \\
\hline
ECG (600)                & 0.628          \\
ETTm2                    & 0.592       \\
Exchange Rate            & 0.441        \\
ETTh2                    & 0.127          \\
Traffic                  & 0.100         \\
Electricity              & 0.089        \\
Air Quality              & 0.009       \\
Weather                  & -0.093    \\
ECG                      & -0.487       \\
ETTm1                    & -0.522      \\
ETTh1                    & -0.627    \\
EEG                      & -0.834   \\
ILI                      & -0.963     \\
Chorales                 & -0.968   \\
\hline
\end{tabular}
\end{table}



\subsection{Interpretability Results}
\begin{figure*}[h!]
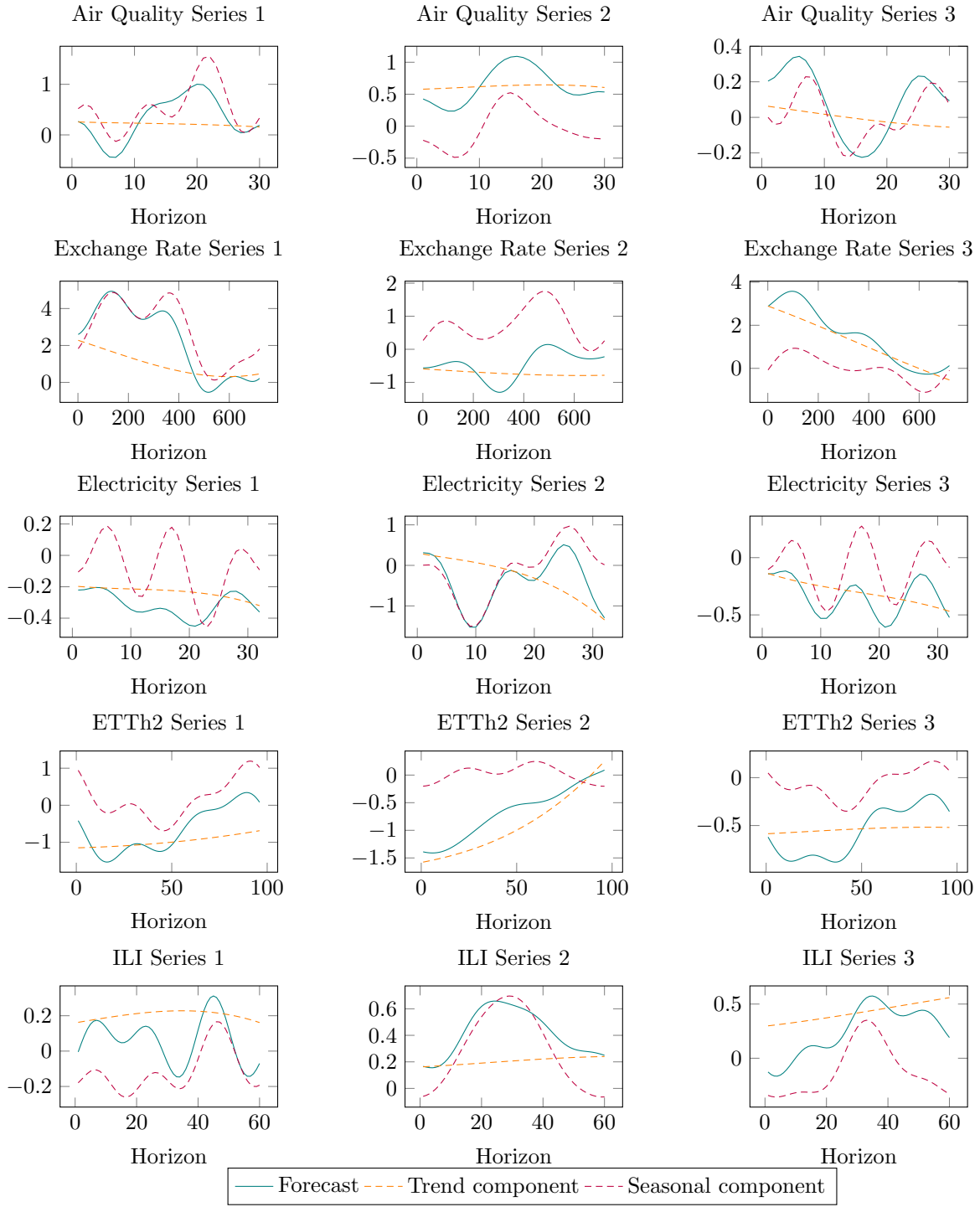

\centering

\caption{Seasonal and Trend components of forecast}
\label{fig:interpretability}
\end{figure*}

Within the NFT model architecture, the breakdown of forecasts into seasonal and trend components is straightforward. This can be achieved by separately considering the outputs of the seasonality and trend blocks. 
Figure~\ref{fig:interpretability} presents the NFT forecasts of the Air Quality, Exchange Rate, Electricity, ETTh2, and ILI datasets for three of their variables.
The trend output exhibits a consistent, gradual progression, while the seasonality output showcases regular cyclic patterns with recurring fluctuations. Notably, in the presence of substantial seasonality in the time series, the seasonality component often exhibits a significantly larger peak-to-peak magnitude compared to the trend component.

\subsection{Ablation Study}
\label{sec:lstm-fc-coefficients}
In this ablation study, we examined the effects of substituting the Temporal Convolutional Network (TCN) layers in our original model with a combination of Long Short-Term Memory (LSTM) layers and Fully Connected (FC) layers, similar to the structure used in the N-BEATS model. The modified LSTM-based architecture has two layers, each with 50 hidden units while maintaining the same input configuration as the original TCN-based setup. The FC variant had four layers, each with 256 units, like the approach in N-BEATS. The training process was similar to the original NFT model.

\begin{figure*}[htbp]
\centering
\begin{tikzpicture}
  \begin{groupplot}[
    group style={
      group size=3 by 4, 
      horizontal sep=2cm,
      vertical sep=1.8cm, 
    },
    width=5cm,
    height=3.5cm,
    legend style={at={(-1,-0.5)}, anchor=north, legend columns=-1}
  ]

  \pgfplotsset{
    nft/.style={blue, solid},
    lstm nft/.style={cyan, dashed},
    fc nft/.style={purple, dashdotted},
  }
    
\nextgroupplot[xlabel={Horizon}, ylabel={MSE}, title={\textbf{Weather}}]
    \addplot[nft] coordinates { (1, 0.19) (7, 0.24) (15, 0.24) (30, 0.25) (60, 0.26) (90, 0.27) (120, 0.28) (180, 0.27) };
\addplot[lstm nft] coordinates { (1, 0.29) (7, 0.33) (15, 0.36) (30, 0.37) (60, 0.39) (90, 0.41) (120, 0.42) (180, 0.44)};
\addplot[fc nft] coordinates { (1, 0.41) (7, 0.39) (15, 0.40) (30, 0.39) (60, 0.40) (90, 0.41) (120, 0.42) (180, 0.41)};

\nextgroupplot[xlabel={Horizon}, ylabel={MSE}, title={\textbf{Air Quality}}]
    \addplot[nft] coordinates { (1, 0.13) (5, 0.21) (10, 0.28) (15, 0.31) (20, 0.35) (25, 0.38) (30, 0.4) };
\addplot[lstm nft] coordinates { (1, 0.38) (5, 1.21) (10, 1.29) (15, 1.25) (20, 1.76) (25, 2.02) (30, 1.78) };
\addplot[fc nft] coordinates { (1, 0.49) (5, 1.04) (10, 1.27) (15, 1.51) (20, 1.62) (25, 1.72) (30, 1.88) };

\nextgroupplot[xlabel={Horizon}, ylabel={MSE}, title={\textbf{ECG (600 patients)}}]
\addplot[nft] coordinates {(1,0.02) (5,0.08) (10,0.07) (15,0.1) (20,0.15) (25,0.15) (30,0.1)};
\addplot[lstm nft] coordinates { (1, 0.07) (5, 0.07) (10, 0.10) (15, 0.14) (20, 0.15) (25, 0.18) (30, 0.21) };
\addplot[fc nft] coordinates { (1, 0.08) (5, 0.08) (10, 0.12) (15, 0.16) (20, 0.19) (25, 0.22) (30, 0.24) };

\nextgroupplot[xlabel={Horizon}, ylabel={MSE}, title={\textbf{ECG}}]
\addplot[nft] coordinates {(1,0.04) (10,0.08) (25,0.12) (50,0.14) (100,0.18) };
\addplot[lstm nft] coordinates { (1, 0.085) (10, 0.3675) (25, 0.4375) (50, 0.5225) (100, 0.5825) };
\addplot[fc nft] coordinates { (1, 0.225) (10, 0.3375) (25, 0.4025) (50, 0.475) (100, 0.5175) };

\nextgroupplot[xlabel={Horizon}, ylabel={MSE}, title={\textbf{EEG}}]
\addplot[nft] coordinates { (1, 0.03) (10, 0.1) (25, 0.16)};
\addplot[lstm nft] coordinates { (1, 0.1) (10, 0.445) (25, 0.655) };
\addplot[fc nft] coordinates { (1, 0.125) (10, 0.425) (25, 0.545) };

\nextgroupplot[xlabel={Horizon}, ylabel={MSE}, title={\textbf{Chorales}}]
 \addplot[nft] coordinates {(1,0.15) (2,0.18) (3,0.20) (4,0.23) (5,0.22)};
\addplot[lstm nft] coordinates { (1, 0.36) (2, 0.3) (3, 0.26) (4, 0.245) (5, 0.23) };
\addplot[fc nft] coordinates { (1, 0.35) (2, 0.29) (3, 0.28) (4, 0.29) (5, 0.28) };

  \addlegendentry{NFT}
  \addlegendentry{NFT (LSTM Layers)}
  \addlegendentry{NFT (FC Layers)}
  \end{groupplot}
\end{tikzpicture}
\caption{MSE results of NFT and NFT with LSTM and FC layers}
\label{fig:NFT_LSTM_MSE_metric}
\end{figure*}
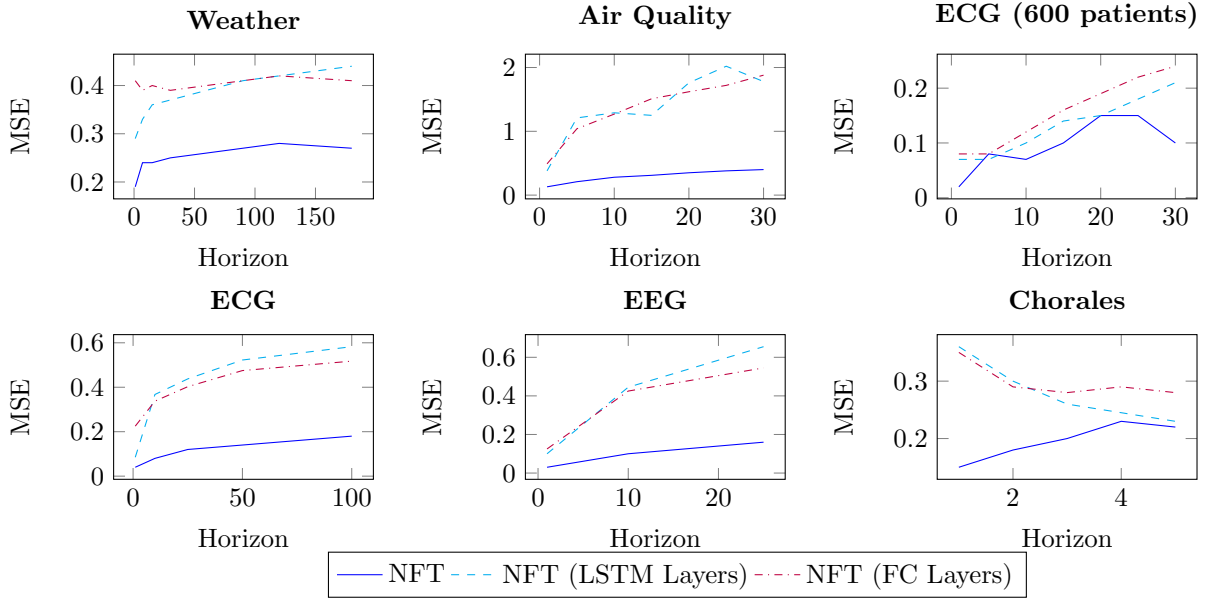

We evaluated the performance of these LSTM-based and FC-based models against the original TCN-based NFT model using the MSE metric. The results are illustrated in Figure \ref{fig:NFT_LSTM_MSE_metric}. 

The NFT model outperformed the LSTM and FC based variants across all datasets. 

For the LSTM-based model, the NFT model reduced the MSE values with average percentages of \textbf{33.37\%} in Weather, \textbf{77.25\%} in Air Quality, \textbf{26.39\%} in ECG (600 patients), \textbf{69.20\%} in ECG, \textbf{74.36\%} in EEG, and \textbf{26.37\%} in Chorales.

For the FC-based model, the NFT model reduced the MSE values with average percentages of \textbf{38.08\%} in Weather, \textbf{77.96\%} in Air Quality, \textbf{37.91\%} in ECG (600 patients), \textbf{72.88\%} in ECG, \textbf{74.37\%} in EEG, and \textbf{33.15\%} in Chorales.




These findings are aligned with our expectations, empirically demonstrating the algorithmic advantage of TCN for time series forecasting.

\section{Conclusions}
In this study, we introduced the Neural Fourier Transform (NFT), a novel approach for multivariate time series forecasting that combines multi-dimensional Fourier transforms with Temporal Convolutional Network layers. We showed that NFT excels in capturing complex seasonal patterns and inter-variable dependencies, offering enhanced precision and interpretability. 
The NFT's utilization of multi-dimensional Discrete Fourier Transform, executed through dual 1-DFT operations, enables a comprehensive exploration of time series data across both temporal and variable dimensions. This approach effectively captures essential patterns, enhancing the model's ability to discern complex temporal dependencies. The integration of TCN layers for learning Fourier coefficients further improves the NFT's forecasting capabilities. Additionally, the NFT is quick and runs efficiently, making it a practical choice for real-world applications.
Tested across fourteen diverse datasets including Electricity, ILI, Traffic, Exchange Rate, ETTh1, ETTh2, ETTm1, ETTm2, ECG, ECG (600 patients), EEG, Air Quality, Bach Chorales, and Weather records, the NFT demonstrated superior predictive accuracy and interpretability over traditional methods across various horizons and lookbacks.
 
In this study, the Neural Fourier Transform (NFT) model was based on the methodology of N-BEATS and its architecture. Looking ahead, an area for investigation is the integration of NFT with other modern forecasting frameworks like Prophet \cite{zunic2020application} and Neural Prophet \cite{triebe2021neuralprophet}. These models, as of now, lack the capability for simultaneous multivariate time series prediction.
Additionally, adapting the NFT model for long-term time series forecasting (LTSF) could significantly enhance its utility in scenarios requiring extended future predictions.
\bibliographystyle{unsrt} 
\bibliography{references} 

\appendix

\end{document}